%% file: main.tex
\documentclass[10pt]{article} 

\usepackage[preprint]{rlj} 

\usepackage{amssymb}            
\usepackage{mathtools}          
\usepackage{mathrsfs}           
\usepackage{graphicx}           
\usepackage{subcaption}         
\usepackage[space]{grffile}     
\usepackage{url}                
\usepackage{lipsum}             
\usepackage{wrapfig}
\usepackage{amsfonts} 
\usepackage{booktabs}
\usepackage{array}
\input{math_commands.tex}

\newcommand{\lp}{\left(}
\newcommand{\ep}{\right)}

\newtheorem{theorem}{Theorem}

\newtheorem{definition}{Definition}[section]

\makeatletter
\newcommand*{\inlineequation}[2][]{%
  \begingroup
    \refstepcounter{equation}%
    \ifx\\#1\\%
    \else
      \label{#1}%
    \fi
    \relpenalty=10000 %
    \binoppenalty=10000 %
    \ensuremath{%
      #2%
    }%
    ~\@eqnnum
  \endgroup
}
\makeatother

\title{PEnGUiN: Partially Equivariant Graph NeUral Networks for Sample Efficient MARL}

\setrunningtitle{PEnGUiN: Partially Equivariant Graph NeUral Networks for Sample Efficient MARL}

\author{Joshua McClellan\textsuperscript{1,2}, Greyson Brothers\textsuperscript{2}, Furong Huang\textsuperscript{1}, Pratap Tokekar\textsuperscript{1}}

\emails{joshmccl@umd.edu, greyson.brothers.jhuapl.edu, furongh@umd.edu, tokekar@umd.edu }

\affiliations{
$^{1}$\textbf{Department of Computer Science, University of Maryland College Park}\\
$^{2}$\textbf{The Johns Hopkins Applied Physics Lab}\\
}

\contribution{
    We present the first generalization of Equivariant Graph Neural Networks (EGNN) to Partial Equivariance with our novel neural network architecture Partially Equivariant Graph Neural Networks (PEnGUiN). We show theoretically that PEnGUiN unifies fully equivariant (EGNN) and non-equivariant (GNN) representations within a single architecture, controlled by a learnable parameter called the symmetry score.
    }
    {
    PEnGUiN builds on EGNN \citep{egnn} and E2GN2 \citep{E2GN2}, and is designed to handle environments with asymmetries, unlike prior work that primarily focuses on full equivariance. 
    }
    \contribution{We show the first Partially Equivariant Neural Network applied to Multi-Agent Reinforcement Learning, leading to improved performance over GNNs and EGNNs in MARL. 
    }
    {
    Prior work has applied equivariance to MARL \citep{eqmmdp, E2GN2}, these approaches typically assume full equivariance. 
    }
    \contribution{
    We formally define and categorize several types of partial equivariance relevant to Multi-Agent Reinforcement Learning (MARL), including subgroup equivariance, feature-wise equivariance, subspace equivariance, and approximate equivariance.
    }
    {
    While specific instances of broken symmetries have been discussed \citep{subequiv_rl, approx_eq_rl}, our work provides a unified and comprehensive categorization tailored to MARL.
    }
\contribution{
    Through experiments on Multi-Particle Environments (MPE) and the highway-env benchmark, we empirically validate that PEnGUiN outperforms both EGNNs and standard GNNs in MARL tasks with various types of asymmetries.
}
{
None
}


\summary{Equivariant Graph Neural Networks (EGNNs) excel at Multi-Agent Reinforcement Learning (MARL) problems by harnessing symmetries in observations, but struggle in real-world environments where symmetries may be broken to varying degrees. We introduce \textit{Partially Equivariant Graph Neural Networks (PEnGUiN)}, a novel architecture that learns to exploit partial symmetries. PEnGUiN blends equivariant and non-equivariant updates via a learnable parameter, adapting to the degree and type of symmetry present and bridging the gap between fully equivariant and non-equivariant models. In addition, we formalize types of partial equivariance common to real-world environments (subgroup, feature-wise, subspace, and approximate). Experiments on MARL benchmarks demonstrate PEnGUiN's superior performance and robustness compared to EGNNs and GNNs in asymmetric settings. PEnGUiN learns where equivariance holds, improving applicability to real-world MARL problems. 
}

\begin{document}

\maketitle  

\begin{abstract}
Equivariant Graph Neural Networks (EGNNs) have emerged as a promising approach in Multi-Agent Reinforcement Learning (MARL), leveraging symmetry guarantees to greatly improve sample efficiency and generalization. However, real-world environments often exhibit inherent asymmetries arising from factors such as external forces, measurement inaccuracies, or intrinsic system biases. This paper introduces \textit{Partially Equivariant Graph NeUral Networks (PEnGUiN)}, a novel architecture specifically designed to address these challenges. We formally identify and categorize various types of partial equivariance relevant to MARL, including subgroup equivariance, feature-wise equivariance, regional equivariance, and approximate equivariance. We theoretically demonstrate that PEnGUiN is capable of learning both fully equivariant (EGNN) and non-equivariant (GNN) representations within a unified framework. Through extensive experiments on a range of MARL problems incorporating various asymmetries, we empirically validate the efficacy of PEnGUiN. Our results consistently demonstrate that PEnGUiN outperforms both EGNNs and standard GNNs in asymmetric environments, highlighting their potential to improve the robustness and applicability of graph-based MARL algorithms in real-world scenarios.
\end{abstract}

\section{Introduction}
\label{sec:introduction}



Multi-Agent Reinforcement Learning (MARL) presents significant challenges due to the complexities of agent interactions, non-stationary environments, \begin{wrapfigure}{htbp}{0.35\textwidth}
    \includegraphics[width=0.35\textwidth]{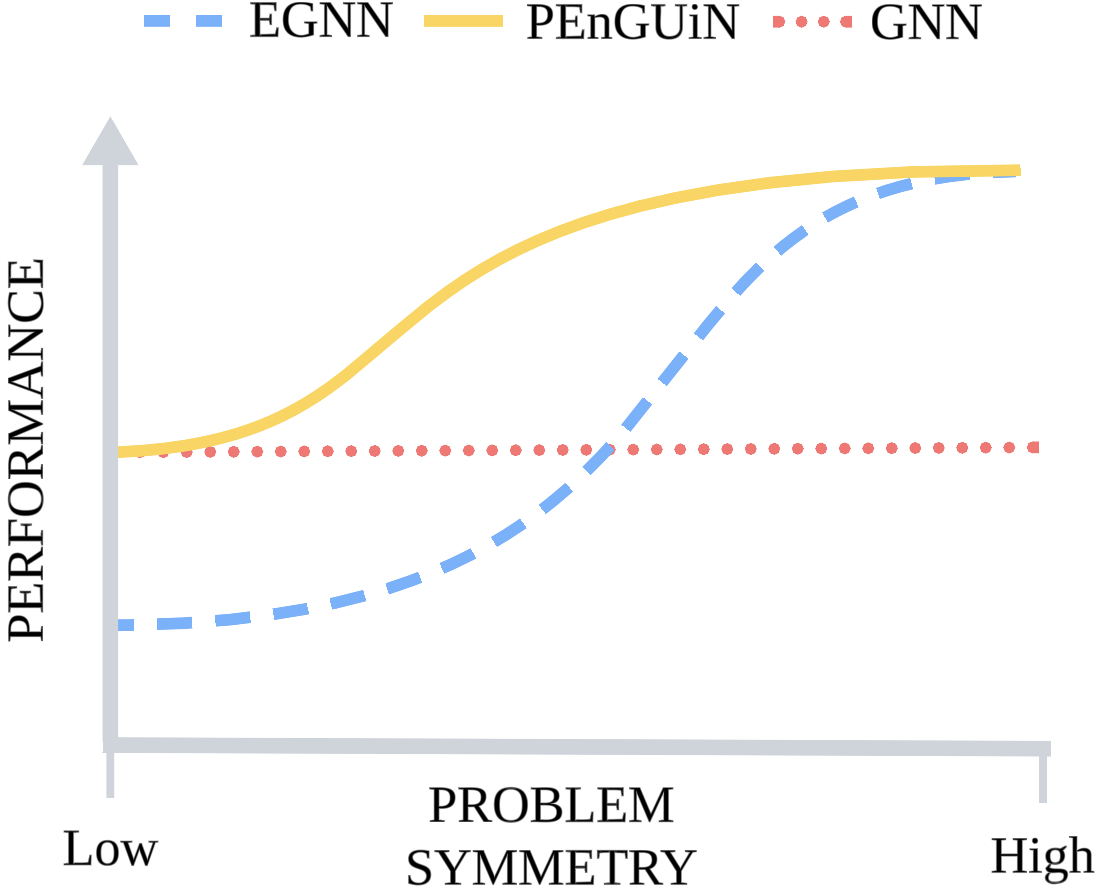}
    \caption{\small{An example of how EGNNs can be advantageous in equivariant environments, and a liability when an environment has increased asymmetries. } }
    \label{fig:tradeoff}
    \vspace{-30pt}
\end{wrapfigure}and the need for efficient exploration and generalization. Recently, Equivariant Graph Neural Networks (EGNNs) \citep{egnn} have emerged as a promising approach in MARL, leveraging inherent symmetries  in multi-agent systems to improve sample efficiency and generalization performance \citep{eqmmdp, E2GN2}. By encoding equivariance to transformations like rotations and translations, EGNNs can achieve superior sample efficiency and generalization, particularly in environments where geometric relationships are crucial.


However, many real-world MARL scenarios do not exhibit perfect symmetry, and there are concerns that \textit{this architecture may be too restrictive in its assumptions}. The real world is messy, and it is rare for something to be exactly rotationally equivariant. Factors such as external forces (e.g., wind, gravity), sensor biases, environmental constraints (e.g., obstacles, landmarks, safety zones), or heterogeneous agent capabilities introduce asymmetries that break the assumptions underlying fully equivariant models. Applying standard EGNNs in these partially symmetric environments can lead to suboptimal performance, as the imposed equivariance constraints may not accurately reflect the underlying dynamics. Conversely, standard Graph Neural Networks (GNNs), which lack any inherent equivariance guarantees, may fail to exploit the symmetries that do exist, leading to reduced sample efficiency and weaker generalization.


This paper introduces Partially Equivariant Graph NeUral Networks (PEnGUiN), a novel architecture designed to address the challenges of learning in partially symmetric MARL environments. PEnGUiN provides a flexible and unified framework that seamlessly \textit{integrates both equivariant and non-equivariant representations} within a single model. Unlike traditional approaches that either enforce full equivariance or disregard symmetries entirely, PEnGUiN learns to \textit{adaptively adjust its level of equivariance} based on the input. This is achieved through a blending mechanism controlled by a learnable parameter that modulates the contribution of equivariant and non-equivariant updates within the network.



Prior works have explored symmetry-breaking cases broadly under the label of ``approximately equivariant'' \cite{approx_eq_def}. This work introduces several more precise categories of partial symmetry that commonly emerge in MARL environments. This includes \textit{subgroup equivariance}, \textit{regional equivariance}, \textit{feature-wise equivariance}, and \textit{general approximate equivariance}. These categories are used to design and test on partially equivariant experiments in the Multi-Particle Environments (MPE) \citep{maddpg} and Highway-env \cite{highway-env}.   Our contributions can be summarized as follows:
    
(1) We present the first generalization of Equivariant Graph Neural Networks (EGNN) to Partial Equivariance with our novel neural network architecture Partially Equivariant Graph Neural Networks (PEnGUiN).  We show theoretically that PEnGUiN unifies fully equivariant (EGNN) and non-equivariant (GNN) representations within a single architecture

(2) We formally define and categorize several types of partial equivariance relevant to Multi-Agent Reinforcement Learning (MARL).

(3) We demonstrate the first Partially Equivariant Neural Network applied to Multi-Agent Reinforcement Learning, leading to improved performance over GNNs and EGNNs in asymmetric MARL. 

\section{Related Works}
Research in equivariant neural networks has explored various architectures and applications, aiming to improve learning and generalization by leveraging symmetries.  Equivariant Graph Neural Networks (EGNNs) \citep{egnn}, SEGNNs \citep{segnn}, and E3NNs \citep{e3nn} are prominent examples, designed to be equivariant to rotations, translations, and reflections. PEnGUiN builds on the EGNN architecture \citep{egnn}, but extends its capabilities to handle partial equivariance.   \citep{Finzi2021} introduced Equivariant MLPs, which are versatile but computationally expensive.  Within reinforcement learning, \citet{eqmdp} and \citet{eqmmdp} established theoretical frameworks for equivariant Markov Decision Processes (MDPs) and Multi-Agent MDPs (MMDPs), respectively, focusing on fully equivariant settings with simple dynamics.  \citet{E2GN2} introduced E2GN2 to address exploration challenges in EGNN-based MARL.  \citet{e3_coop} employed SEGNNs for cooperative MARL, though SEGNNs often have slower training times.  \citet{rl_symloss} explored adding a symmetry-based loss term, showing limited performance gains. \citet{Wang2022} investigated rotation equivariance for robotic manipulation with image-based observations.  These works primarily address \textit{full} equivariance, or focus on specific tasks or symmetry types, contrasting with PEnGUiN's general and learnable approach to \textit{partial} equivariance.

Research on partial or approximate equivariance includes group CNNs for image processing \citep{approx_eq_def, relax_groupConv, almost_eq_lieConv, matrix_cnn_approx, relax_eq_filters, approx_eq_rl} and combining MLPs with equivariant components \citep{finzi_residual_2021}, which are distinct from our graph-based approach. Studies \citep{eff_latent_sym, extrinsic_eq_theory, approx_gen_group_eq} have analyzed the effectiveness of equivariant models in asymmetric scenarios, motivating models like PEnGUiN that can learn equivariance quantities. In the realm of GNNs, \citet{relax_eq_gnn} concurrently introduce a relaxed equivariant GNN; however, their model is built upon spherical harmonic representations (which increases implementation and computation complexity), unlike PEnGUiN, which is based on EGNNs and allows a smooth transition between fully equivariant and standard GNN behavior. \citet{villar_approx_gnn} studies GNN permutation equivariance, not O(n) equivariance. \citet{subequiv_rl} developed subgroup equivariant GNNs tailored for robotics, specifically to ignore gravity, limiting their applicability compared to PEnGUiN's general framework.

\section{Background}
\subsection{Multi-Agent Reinforcement Learning} Multi-Agent Reinforcement Learning (MARL) extends the principles of Reinforcement Learning (RL) to scenarios involving multiple interacting agents within a shared environment. In MARL, each agent aims to learn an optimal policy $\pi_i$ that maximizes its own expected cumulative reward $R_i$, which is influenced by the actions of other agents and the environment dynamics.  Formally, at each timestep $t$, each agent $i$ observes a local state $s_i^t$, takes an action $a_i^t$ according to its policy $\pi_i(a_i^t | s_i^t)$, and receives a reward $r_i^t = R_i(s^t, a^t)$, where $s^t = (s_1^t, ..., s_N^t)$ and $a^t = (a_1^t, ..., a_N^t)$ represent the joint state and action spaces of all $N$ agents \citep{littman1994markov}. The goal of each agent $i$ is to learn a policy $\pi_i(a_i|s)$ that maximizes its expected return: $J(\pi_i) = \mathbb{E}{\pi_1,...,\pi_N} \big[ \sum_{t=0}^T \gamma^t R_i(s_t, a^1_t,...,a^N_t) \big]$ where $T$ is the time horizon, $\gamma \in (0,1]$ is a discount factor, and $a^j_t \sim \pi_j(\cdot|s_t)$.

\begin{figure*}[t]
\centering
\begin{minipage}{0.204\textwidth}
  \centering
\includegraphics[width=1.0\textwidth]{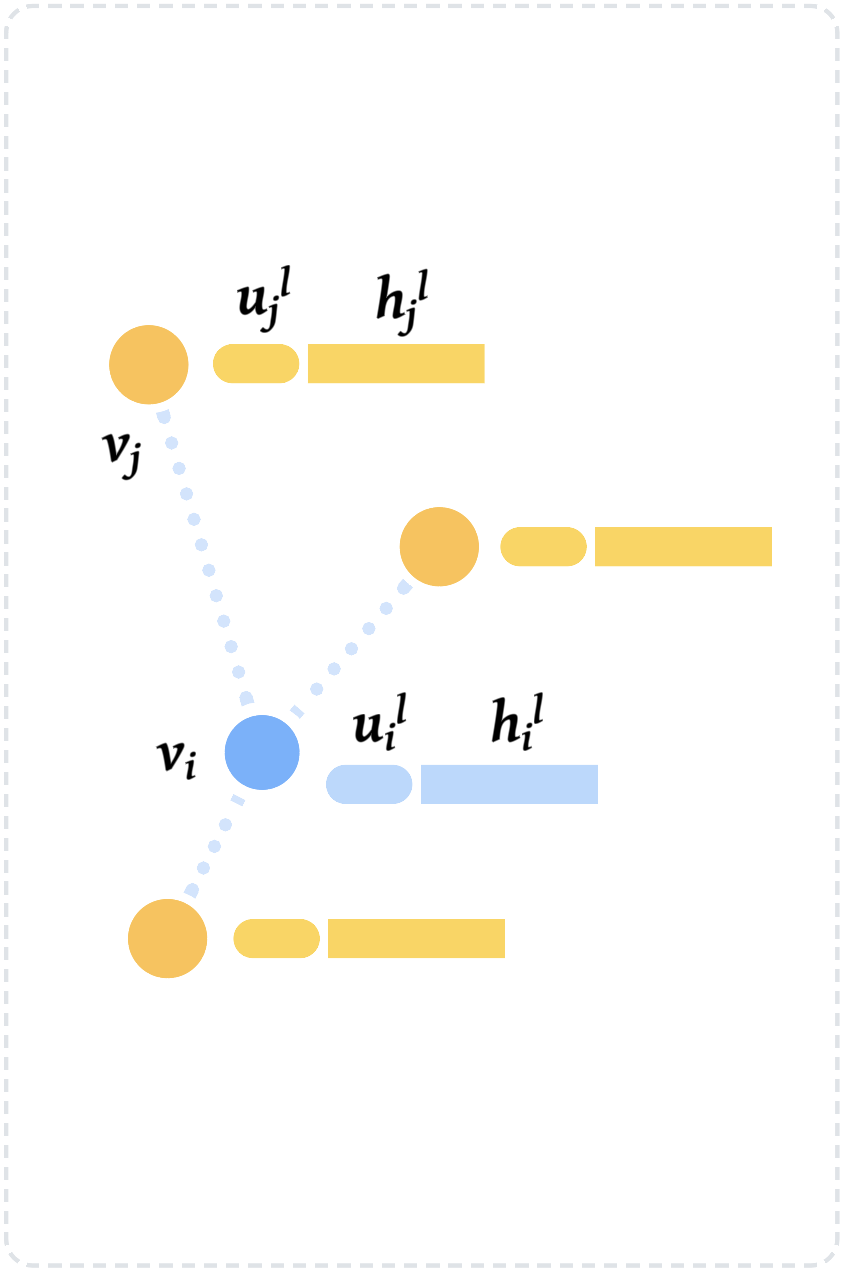}
\subcaption[first caption.]{Graph at layer $l$}\label{fig:layer-graph}
\end{minipage}%
\hspace{0.15ex}
\begin{minipage}{0.43\textwidth}
  \centering
\includegraphics[width=1.0\textwidth]{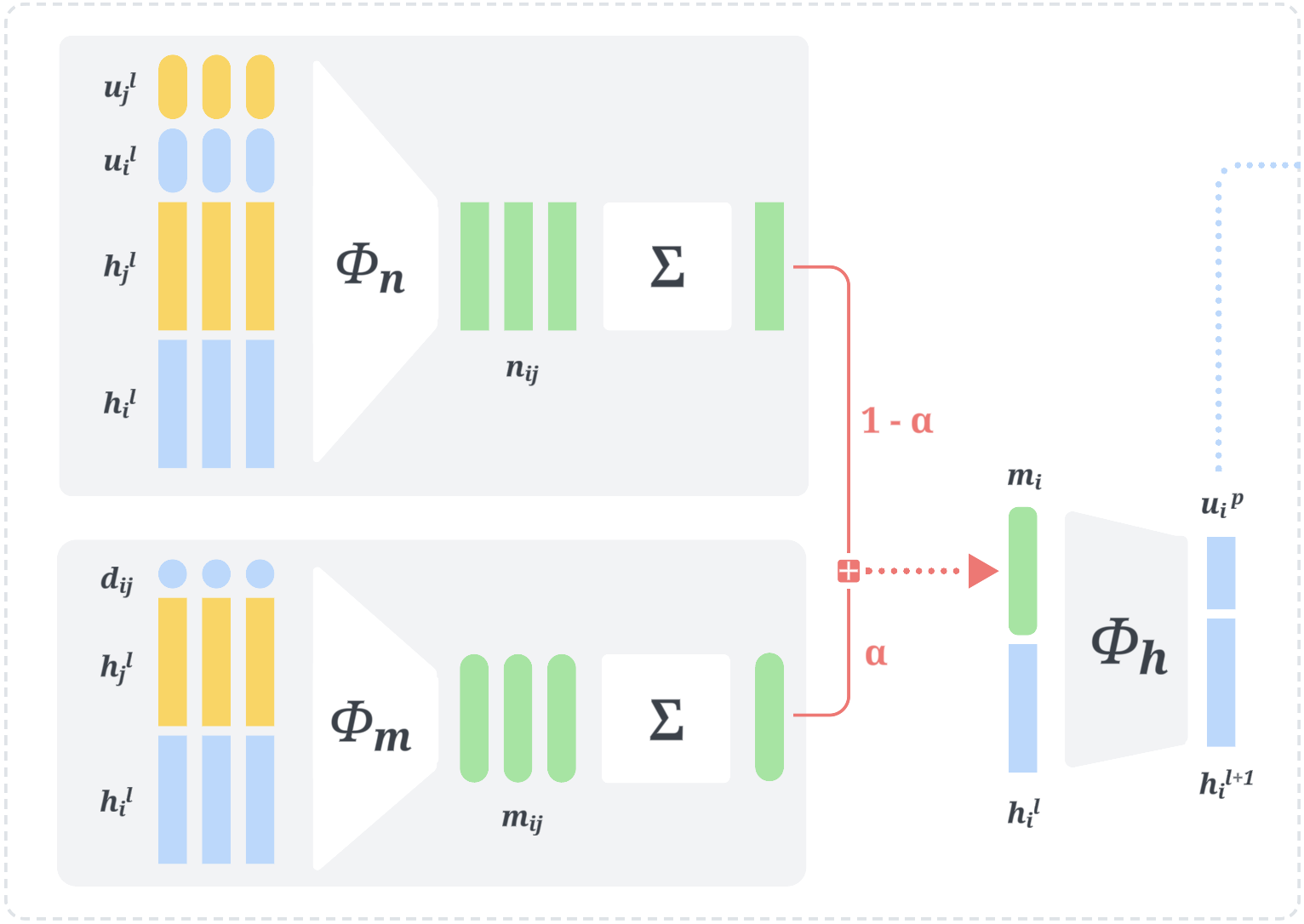}
\subcaption[first caption.]{Node Feature Update $\ \vh_i^{l+1}$}\label{fig:layer-feature}
\end{minipage}%
\hspace{0.15ex}
\begin{minipage}{0.336\textwidth}
  \centering
\includegraphics[width=1.0\textwidth]{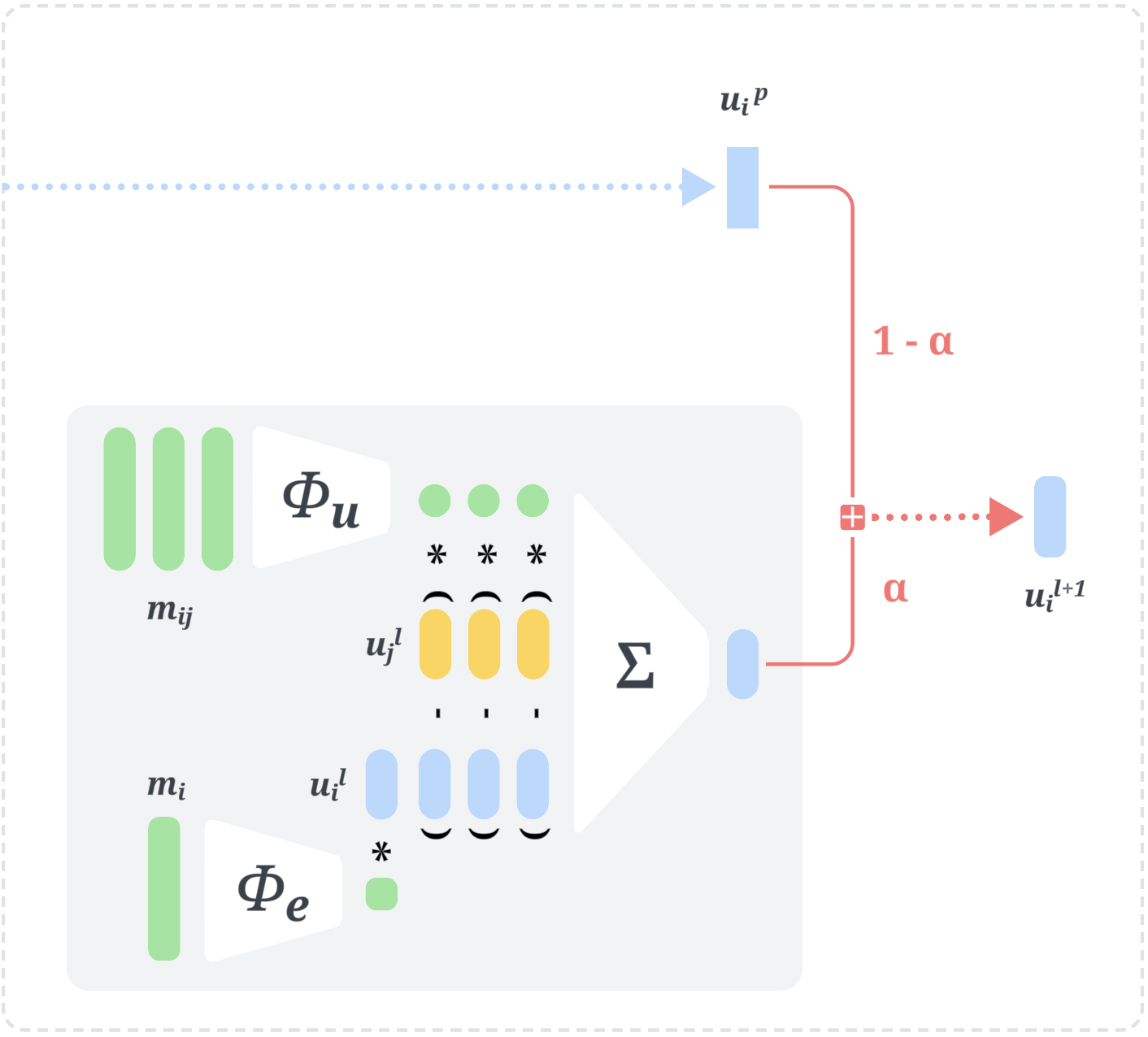}
\subcaption[second caption.]{Node Coordinate Update $\ \vu_i^{l+1}$}\label{fig:layer-coordinate}
\end{minipage}%
\caption{\small{Diagram of an individual PEnGUiN layer described in section \ref{sec:penguin}. The colored boxes represent vectors, where rounded corners indicate the preservation of equivariance and square corners indicate non-equivariance. An example graph is provided in (a), showing coordinates $\vu_i$ and features $\vh_i$ corresponding to each node $v_i$. The update for node $v_i$ (blue) is split into feature and coordinate updates, shown in (b) and (c) respectively. Within each subfigure is a non-equivariant branch (top) and an equivariant branch (bottom), whose outputs are blended via convex combination (red) governed by the symmetry score $\alpha$. } }
\label{fig:penguin-diagram}
\end{figure*}

\subsection{Equivariance} Equivariance describes how functions behave under transformations. A function $f$ is said to be equivariant to a group of transformations $G$ if transforming the input $x$ by a group element $g \in G$ results in a predictable transformation of the output $f(x)$. Formally, if $T_g$ represents a transformation of the input space and $L_g$ represents a transformation of the output space, equivariance is defined as:$ f(T_g x) = L_g f(x), \quad \forall g \in G, \forall x $. Related to equivaraince is invariance, where the output remains unchanged under the input transformation, i.e., $f(T_g x) = f(x)$.

\section{Partially Equivariant Graph Neural Networks}
\label{sec:penguin}
To address the challenges of learning in partially symmetric environments, we introduce Partially Equivariant Graph Neural Networks (PEnGUiN). PEnGUiN is a novel graph neural network architecture designed to seamlessly incorporate varying degrees of equivariance, ranging from full $O(n)$ equivariance, as in E2GN2s, to non-equivariant behavior, akin to standard GNNs.  This flexibility is achieved through a blending mechanism controlled by a parameter $\alpha$, allowing the network to adapt to and learn the specific symmetries present in the data. PEnGUiN follows a similar message-passing paradigm as a standard GNN with message computation, message aggregation, and node feature updates. The forward pass of a single layer $l$ in PEnGUiN, shown in Figure \ref{fig:penguin-diagram}, is defined by the following equations:

\begin{table}[htbp]
  \centering
  \caption{PEnGUiN Update Equations for layer $l$}
  \label{tab:penguin_equations_compact}
  \begin{tabular}{p{15em}p{22em}}
    \toprule
    \textbf{Message Computation:} & 
    \quad Equivariant: $\vm_{ij}^{l} =\phi_{m}\left(\vh_{i}^{l}, \vh_{j}^{l},  \|\vu_{i}^{l}-\vu_{j}^{l}\|^{2}\right)$ \\
    &\quad Non-equivariant: $\vn_{ij}^{l} =\phi_{n}\left(\vh_{i}^{l}, \vh_{j}^{l},\vu_{i}^{l}, \vu_{j}^{l}\right)$ \\
    \midrule
    \textbf{Message Aggregation:} &
    \quad $\vm_{i}^{l} = \alpha \sum_{j \neq i} \vm_{ij}^l + (1-\alpha) \sum_{j \neq i} \vn_{ij}^l$ \\
    \midrule
        \textbf{Equivariant Coordinate Update:} &
           \quad  $ \vu_{i,eq}^{l} = \vu_{i}^{l} \phi_{e}(\vm_i^l) + \sum_{j \neq i} \left(\vu_{i}^{l} -\vu_{j}^{l}\right) \phi_{u} \left(\vm_{ij}^l\right) $\\
    \midrule
    \textbf{Feature Update:}
    &\quad $\vh_{i}^{l+1}, \vu_i^p =\phi_{h}\left(\vh_{i}^{l}, \vm_{i}^{l}\right)$ \\
    \midrule
    \textbf{Partially Equivariant Coordinate Update:} &
    \quad $\vu_{i}^{l+1} = \alpha  \vu_{i,eq}^l  + (1-\alpha) \vu_i^p$ \\
    \bottomrule
  \end{tabular}
\end{table}

Each node $i$ contains two vectors of information: the node embeddings $\vh_i \in \sR^h$ and the coordinate embeddings $\vu_i \in \sR^n$. The node embeddings are \textit{invariant} to $O(n)$. Inputs for layer $0$ for $\vh_i$ may be information about the node itself, such as node type, ID, or status. The coordinate embeddings for node $i$ are \textit{equivariant} to $O(n)$, and inputs will typically consist of positional values (see the appendix for a discussion on how to incorporate velocity and angles). 

A layer is updated by first computing the non-equivariant $\vn_{ij} \in \sR^m$ and equivariant $\vm_{ij} \in \sR^m$ messages between each pair of nodes $i$ and $j$. Each node then aggregates these messages across all neighboring nodes. At this stage, the aggregated non-equivariant and equivariant messages are mixed together. Finally, the updated feature node vector $\vh_i^{l+1}$  for layer $l+1$ is computed by passing the aggregated message through an MLP $\phi_h:  \sR^{h+m} \mapsto \sR^{h+n}$. This update includes a skip connection to the previous feature node vector. Note that the output of $\phi_h$ is split into $\vh_i \in \sR^m$ and $\vu_i^p \in \sR^n$ (the latter is used in the Partially Equivariant Coordinate update).

The equivariant coordinate vector is updated using the learnable functions (typically MLPs) $\phi_e: \sR^m \mapsto \sR$ and $\phi_u: \sR^m \mapsto \sR$. This update in table \ref{tab:penguin_equations_compact} is guaranteed to be equivariant to $O(n)$ \cite{egnn}. Finally, in the Partially Equivariant update, the equivariant term $\vu_{i,eq}$ is mixed with a non-equivariant component $\vu_{i}^p \in \sR^n$.


A key element of PEnGUiN is the addition of the term $\alpha \in (0,1) \subset \sR$ to quantify the amount of equivariance in the system. For convenience, we will refer to $\alpha$ as the "symmetry score". The value of the symmetry score has the following important implications:

\begin{theorem}
Given a Partially Equivariant Graph Neural Network Layer as defined in table \ref{tab:penguin_equations_compact}, when $\alpha = 1$ the Partial Equivariant Layer is exactly equivalent to an E2GN2 layer. (see Appendix A for proof)
\end{theorem}

An important implication of this theorem is when $\alpha=1$ PEnGUiN is exactly equivariant to rotations (the group $O(n)$). This theorem establishes that PEnGUiN embeds EGNN as a special case. When $\alpha = 1$, PEnGUiN fully exploits the benefits of the equivariant inductive bias, such as improved sample efficiency and generalization in environments with symmetric observations. 

\begin{theorem}
Given a Partially Equivariant Graph Neural Network as defined in table \ref{tab:penguin_equations_compact}, when $\alpha = 0$ the Partial Equivariant Graph Neural Network is equivalent to a GNN (see Appendix A for proof)
\end{theorem}

This theorem highlights PEnGUiN's ability to operate in asymmetric settings.  As $\alpha$ approaches 0, the network's reliance on equivariant updates diminishes, allowing it to learn arbitrary, non-equivariant relationships.

In practice, the amount of equivariance will rarely be a simple constant. Equivariance may be restricted to a certain region, or a subset of features.  Thus, we estimate $\alpha$ using an MLP as a function of the input features for each node: $\phi_{\alpha}(\vh_i^0, \vx_i^0) = \alpha$. We will refer to this network as the Equivariance Estimator (EE). This allows $\alpha$ to be learned as a spatially and entity-dependent function, enabling the network to adaptively modulate equivariance within the network.

\begin{figure*}[t]
\centering
\begin{minipage}{0.23\textwidth}
  \centering
\includegraphics[width=1.0\textwidth]{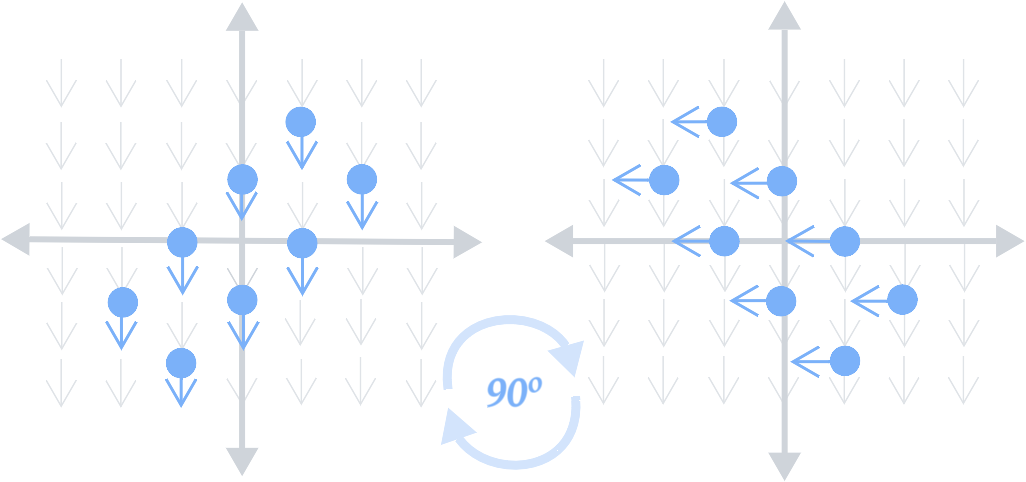}
\subcaption[fourth caption.]{External Forces}\label{fig:sym-dynamics}
\end{minipage}%
\hspace{.02\textwidth}
\begin{minipage}{0.23\textwidth}
  \centering
\includegraphics[width=1.0\textwidth]{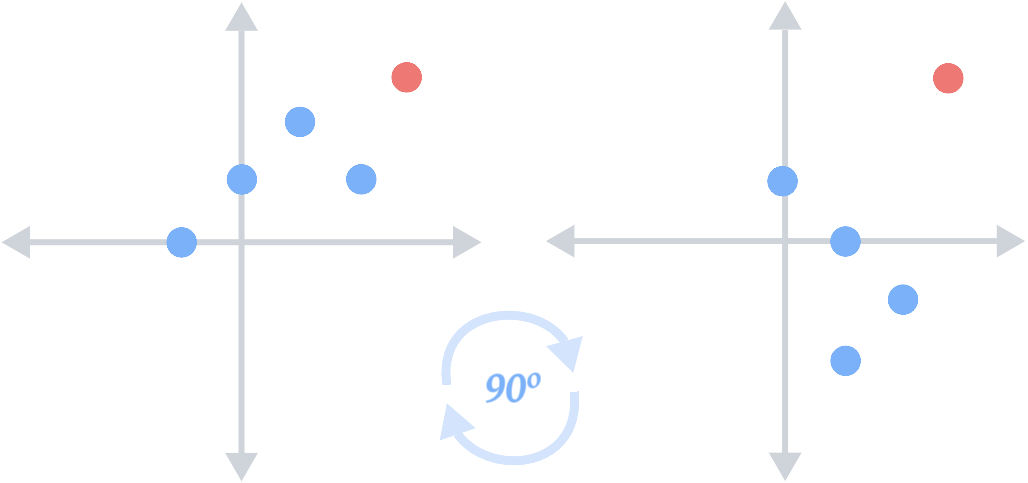}
\subcaption[first caption.]{Fixed Obstacle}\label{fig:sym-decoy}
\end{minipage}%
\hspace{.02\textwidth}
\begin{minipage}{0.23\textwidth}
  \centering
\includegraphics[width=1.0\textwidth]{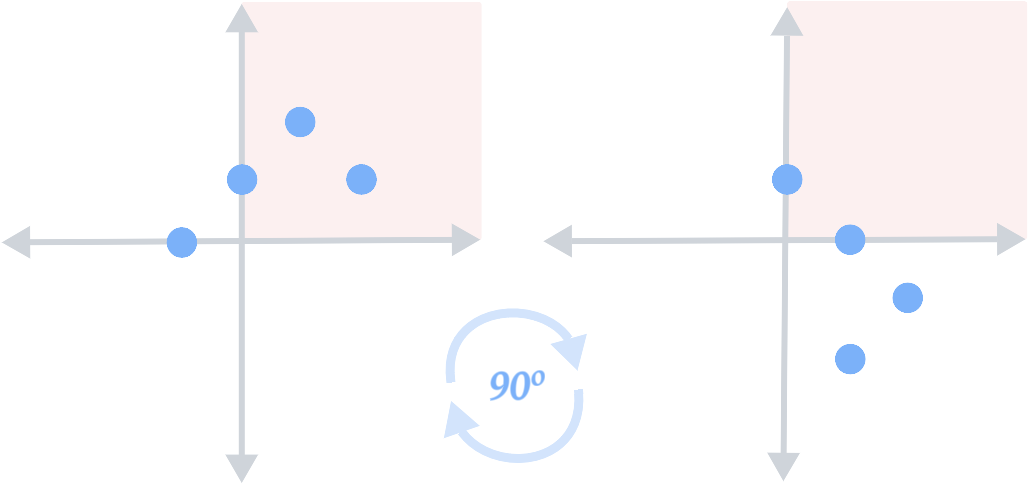}
\subcaption[second caption.]{Safety Region}\label{fig:sym-region}
\end{minipage}%
\hspace{.02\textwidth}
\begin{minipage}{0.23\textwidth}
  \centering
\includegraphics[width=1.0\textwidth]{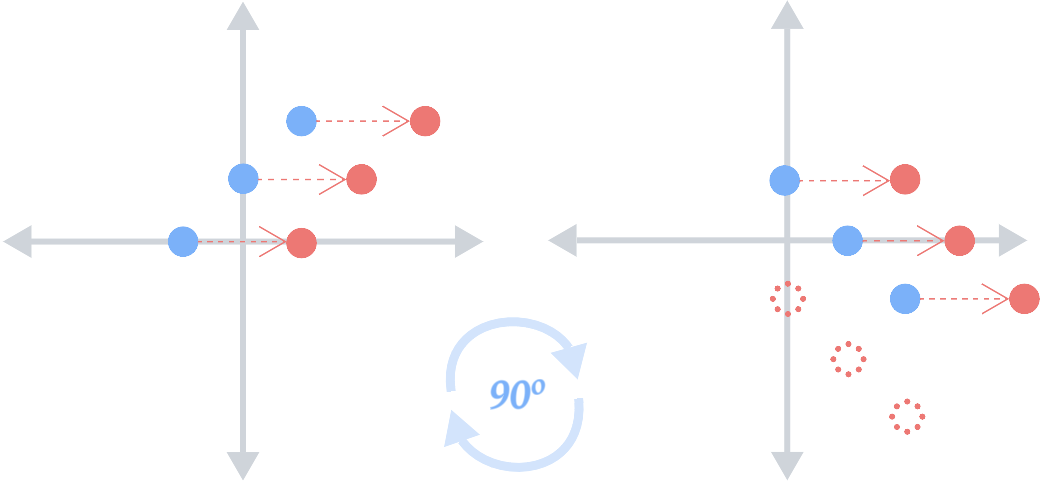}
\subcaption[third caption.]{Sensor Bias}\label{fig:sym-sensor}
\end{minipage}%
\caption{\small{Examples of the types of partial equivariance described in Table \ref{tab:partial_equivariance}, with respect to a 90-degree clockwise rotation about the origin.}} 
\label{fig:partial-equivariance}
\end{figure*}

\section{Categories of Partial Equivariance}

Previous works have noted that functions may have some error in equivariance \citep{approx_eq_def}. Others have noted that functions may be equivariant to subgroups instead of an entire group \citep{subequiv_rl}. In this work, we present a new formalism to unify these asymmetries.  We refer to partial equivariance as any situation with asymmetries.

We divide partial equivariance into four categories: subgroup equivariance, feature-wise equivariance, regional equivariance, and approximate equivariance. Approximate equivariance and subgroup equivariance were previously defined in \citep{approx_eq_def} and \citep{subequiv_rl} respectively. Recall that an equivariant function $f$  will result in the following equality: $\|f(T_g x) - L_g f(x)\| = 0$ where $G$ is a group with a representation tranformation $T_g$ acting on the input space and a representation $L_g$ acting on the output space. 
\begin{table}[htbp]
  \centering
  \caption{\small{Types of Partial Equivariance}}
  \label{tab:partial_equivariance}
  \begin{tabular}{p{10em}cp{10em}}
    \toprule
    \textbf{\small{Type Name}} & \textbf{\small{Equation}} & \textbf{\small{Examples}} \\
    \midrule    
    \textbf{\small{Relaxed/Approximate Equivariance}} &
    \small{$ \| f(T_g x) - L_g f(x)) \| \leq \epsilon$} &
        \small{External forces, nonlinear dynamics, sensor errors.}
    \\    
    \midrule
    \small{\textbf{Subgroup Equivariance}} &
    \small{$f(T_h x) = L_h  f(x), \quad \forall h \in H \subseteq G$} &    
    \small{Ignoring the gravity vector.}
    \\
    \midrule
    \small{\textbf{Feature-Wise Equivariance}} &
    \small{$f(T_g x_1, x_2) = L_g  f(x_1, x_2)$} &
    \small{Fixed Obstacles.}
    \\
    \midrule
    \small{\textbf{Regional Equivariance}} &
    \small{$\|f(T_g x) - L_g f(x)\| = \epsilon(x)$}
     &
        \small{Safety regions.}
    \\
 
    \bottomrule
  \end{tabular}
\end{table}


\begin{definition}[Approximate Equivariance]
Let $f: \mathcal{X} \rightarrow \mathcal{Y}$ be a function The function $f$ is approximately equivariant if there exists a small constant $\epsilon > 0$ such that: $\| f(T_g x) - L_g f(x) \| \leq \epsilon, \quad \forall x \in \mathcal{X}, \quad \forall g \in G $

\end{definition}

 Approximate equivariance is the most general category of Partial Equivariance. Approximate equivariance means that the function is almost equivariant, but there might be small deviations from perfect equivariance.   This is a relaxation of the strict equality required for perfect equivariance. Multi-agent systems with unpredictable wind, nonlinear dynamics, or sensor errors may result in approximate equivariance.


\begin{definition}[Subgroup Equivariance]
  A function $f: \mathcal{X} \rightarrow \mathcal{Y}$ is subgroup equivariant with respect to a subgroup $H \subseteq G$ if, for all $h \in H$ and all $x \in \mathcal{X}$ the following is true $ f(T_h x) = L_h f(x) $
\end{definition}
As an example of subgroup equivariance, consider a quadcopter operating in 3d space. Previous works have shown this will not be equivariant in $E(3)$, specifically due to the effects of the gravity vector (i.e. rotating in the x-z plane affects the dynamics). Instead, \citep{subequiv_rl} only enforced equivariance to the group orthogonal to the gravity vector, that is the subgroup of $E(3)$ that only includes rotations orthogonal to gravity.


\begin{definition}[Feature-wise equivariance]
Let $x = (x_1, x_2, ..., x_n)$ be an input vector where each $x_i$ represents a different feature or subset of features.    A function $f$ is feature-wise equivariant if:  $f(T_g x_1, x_2, ..., x_n) = L_g f_1(x), f_2(x), ..., f_m(x)$ Where $f(x) = (f_1(x), f_2(x), ..., f_m(x))$. 
\end{definition}
Feature-wise equivariance applies when only part of the input is subject to a symmetry transformation. The function is equivariant with respect to that part of the input, while other parts might be invariant or behave in a non-equivariant way.  This allows us to handle situations where some entities of the environment are symmetric, and others are not.



\begin{definition}[Regional Equivariance]
 Let $f: \mathcal{X} \rightarrow \mathcal{Y}$ be a function,  The function $f$ is regional equivariant if there exists a subspace $\mathcal{S} \subset \mathcal{X}$ such that for all $x \in \mathcal{S}$:

$$ \epsilon(x) =  \|f(T_g x) - L_g f(x)\|, \quad \quad \forall g \in G$$
where $\epsilon(x) > 0  \quad if \quad x \in S, \quad and \quad \epsilon(x) = 0 \quad if \quad  x \notin  \mathcal{S}$
\end{definition}
Regional equivariance means that the function exhibits perfect equivariance \textit{only within a specific region or regional of the input space}.  Outside this region, the equivariance property might not hold, or it might be violated to varying degrees.

\section{Experiments}

This section presents an empirical evaluation of Partially Equivariant Graph Neural Networks (PEnGUiN) to address the following key questions: (1) \textit{ Does PEnGUiN offer performance improvements over standard Equivariant Graph Neural Network structures (i.e. EGNN, E2GN2)?} (2) \textit{Is PEnGUiN capable of effectively identifying and leveraging symmetries where they exist while accommodating asymmetries where necessary?} (3) \textit{Does the Equivariance Estimator component of PEnGUiN correctly estimate Partial Equivariance?}
To investigate these questions, we conducted experiments on the Multi-Particle Environments (MPE) benchmark suite \citep{maddpg} and the more complex highway-env benchmark \cite{highway-env}. We compared the performance of PEnGUiN against several baselines using the Proximal Policy Optimization (PPO) \citep{ppo} algorithm implementation from RLlib \citep{rllib}. 

\subsection{Multi-Particle Environment (MPE)}

We utilized two representative scenarios from the MPE benchmark \citep{maddpg}. \textbf{Simple Tag:} a classic predator-prey environment where multiple pursuer agents, controlled by the RL policy, aim to collide with a more nimble evader agent controlled by a heuristic policy to evade capture. The environment also includes static landmark entities. \textbf{Simple Spread:} a cooperative environment in which three agents are tasked with positioning themselves over three landmarks. Agents receive a dense reward for being close to landmarks and are penalized for collisions with each other. These MPE scenarios provide a simplified setting to initially assess the capabilities of PEnGUiN in environments with varying degrees of symmetry.
\begin{figure}[t]
    \begin{subfigure}[h]{0.99\textwidth}
    \centering
    \includegraphics[width=0.99\columnwidth]{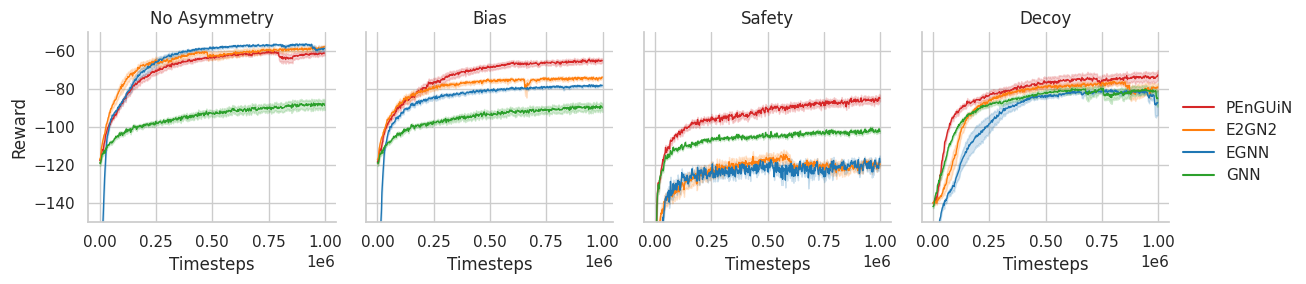}
    \end{subfigure}
    \begin{subfigure}[h]{0.99\textwidth}
    \centering
    \includegraphics[width=0.99\columnwidth]{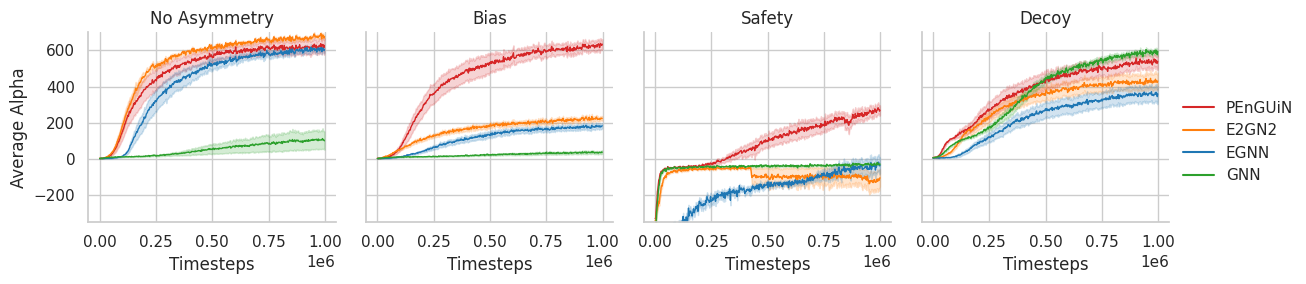}
    \end{subfigure}
    \caption{\small{Learning curves on MPE \textit{simple spread} (Top) and \textit{simple tag} (Bottom) environments under `None', `Bias', and `Safety` asymmetry conditions. Results are averaged over 10 seeds with shaded regions indicating standard error.  PEnGUiN shows consistent performance, especially in environments with feature-wise and regional equivariance.}}
    \vspace{-20pt}
    \label{fig:mpe_learning_curves_top}
\end{figure}
To systematically evaluate PEnGUiN's ability to handle partial equivariance, we introduced three distinct types of asymmetries into the MPE scenarios, corresponding to the categories previously defined:
\begin{enumerate}
    \item \textbf{Sensor Bias (Approximate Equivariance):} We introduced a constant positional bias to the observations of a subset of entities. In \textit{simple tag}, this bias was applied to the observed positions of landmarks and the evader agent. In \textit{simple spread}, the bias was applied to the landmark observations. Critically, this bias was consistently applied to entities not belonging to the agent's team, mimicking biased sensor measurements of external entities.
    \item \textbf{Safety Region (Regional Equivariance):} We implemented a safety region by imposing a negative reward penalty whenever an agent entered the upper-right quadrant of the environment. This creates a spatially defined asymmetry.
    \item \textbf{Decoy (Feature-wise Equivariance):} To test feature-wise equivariance, we added a "decoy" entity to the environment. This decoy visually resembled the agents' objective (evader in \textit{simple tag}, landmarks in \textit{simple spread}) but provided no reward upon interaction.  The true objective remained static, while the decoy moved randomly, introducing an asymmetry based on object identity and reward relevance.
\end{enumerate}

We employed the RLLib PPO agent for training all neural network architectures. We compared PEnGUiN against the following baselines. \textbf{EGNN:} Equivariant Graph Neural Network \cite{egnn}, representing a fully equivariant baseline. \textbf{E2GN2:}  An unbiased version of EGNNs, with improved MARL performances \citep{E2GN2}.  \textbf{GNN:} A standard Graph Neural Network, serving as a non-equivariant baseline.

For PEnGUiN, the $\alpha$ parameter was implemented as a Multi-Layer Perceptron (MLP) that takes as input the node's position $\vu_i^l$ and node type. This allows $\alpha$ to be learned as a spatially and entity-dependent function, enabling the network to adaptively modulate equivariance.

\subsubsection{Results and Discussion (MPE)}
Figure \ref{fig:mpe_learning_curves_top} presents the learning curves for PEnGUiN, EGNN, E2GN2, and GNN across the standard MPE scenarios and their partially equivariant modifications. In the fully symmetric "None" condition, EGNN and E2GN2 achieve strong performance, validating the benefits of equivariance in symmetric environments. However, their performance significantly degrades in the "Bias" and "Safety" scenarios, demonstrating their sensitivity to symmetry breaking.  In contrast, PEnGUiN consistently maintains high performance across all asymmetry conditions, showcasing its robustness and adaptability to partial equivariance. While the standard GNN is less affected by the introduced biases, it consistently underperforms PEnGUiN and equivariant models in symmetric settings, and does not reach the peak performance of PEnGUiN in asymmetric ones. In the "Decoy" environment, PEnGUiN also exhibits superior performance, indicating its effectiveness in handling feature-wise asymmetries.
\begin{wrapfigure}{hr}{0.4\textwidth}
    \includegraphics[width=0.4\textwidth]{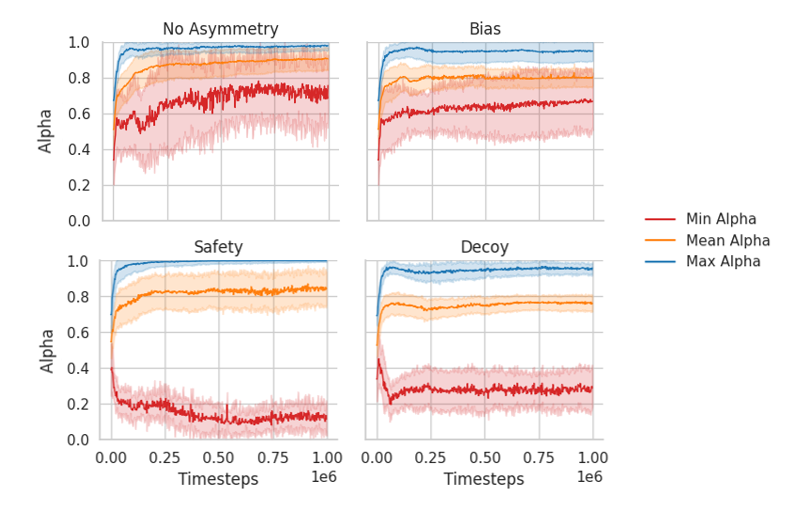}
    \caption{\small{Descriptive statistics of $\alpha$ over training for simple tag. Each statistic is averaged over all 10 seeds for training.} }
    \label{fig:alpha_training}   
    \vspace{-40pt}
\end{wrapfigure}

     

\subsection{PEnGUiN Quantifying Partial Equivariance}
Next, we want to explore how well PEnGUiN identifies Partial Equivariance. In theory, we expect the symmetry score ($\alpha$) to increase as certain regions or features remain equivariant. As asymmetries are introduced into the scenario, the symmetry score should decrease in value where those asymmetries are present.

During training we tracked the average, minimum, and maximum values of the symmetry score. We show these results for the simple tag environment in figure \ref{fig:alpha_training}. \begin{wrapfigure}{l}{0.5\textwidth}
    \vspace{-5pt}
    \begin{subfigure}[h]{0.2\textwidth}
    \centering
    \includegraphics[width=\textwidth]{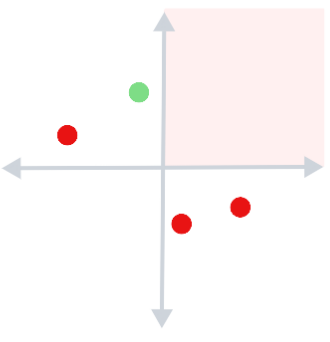}
    \end{subfigure}    
    \begin{subfigure}[h]{0.29\textwidth}
    \centering
    \includegraphics[width=\textwidth]{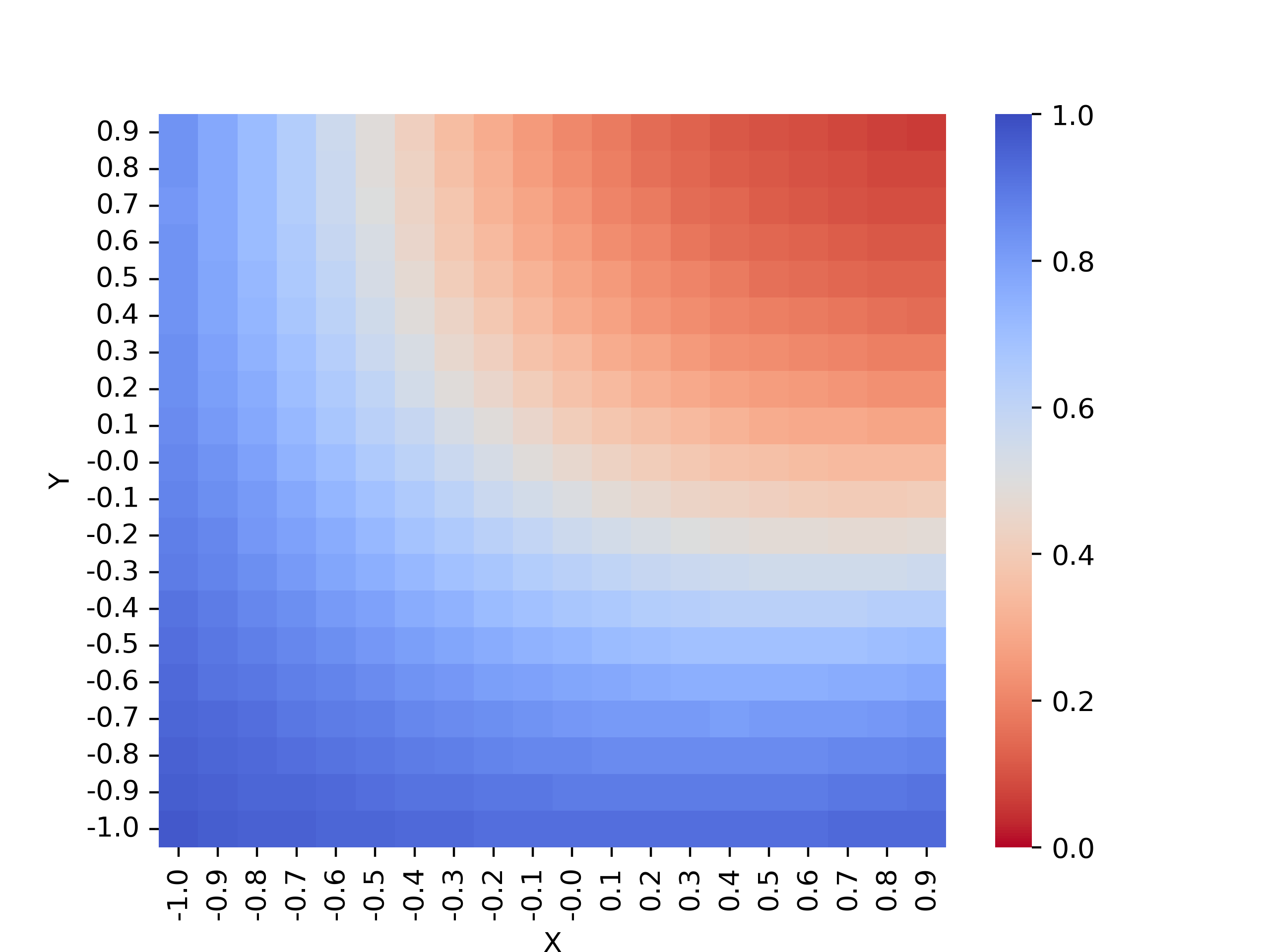}
    \end{subfigure}
    \caption{\small{Left: the 'Safety Region' of the simple tag scenario. Right: visualization of learned $\alpha$ values for PEnGUiN in the "Safety Region" scenario. }}    
    \label{fig:alpha_heatmap} 
    \vspace{-10pt}
\end{wrapfigure} For the scenario with no asymmetries, the symmetry score increases quickly. PEnGUiN is able to learn that the equivariance applies across the scenario. However, it does not reach the exact optimal symmetry score, which would be 1 for this scenario. For the safety and decoy scenarios, we note that the minimum value of $\alpha$ decreases rapidly. It is important to note that the average value seems to stabilize rather quickly, so it appears that learning for the symmetry score occurs primarily in the early stages of training. 

To further investigate PEnGUiN's learned behavior, we visualized the equivariance estimator output in the "Safety Region" scenario (Figure \ref{fig:alpha_heatmap}).

The heatmap shows the output of the equivariance estimator as a function of agent position (X and Y coordinates). Lower $\alpha$ values (red) indicate reduced equivariance, while higher $\alpha$ values (blue) represent stronger equivariance. PEnGUiN learns to reduce equivariance in the designated safety region (upper-right quadrant), effectively adapting to the regional asymmetry.
\begin{wrapfigure}{hr}{0.25\textwidth}
    \vspace{-25pt}
    \includegraphics[width=0.25\textwidth]{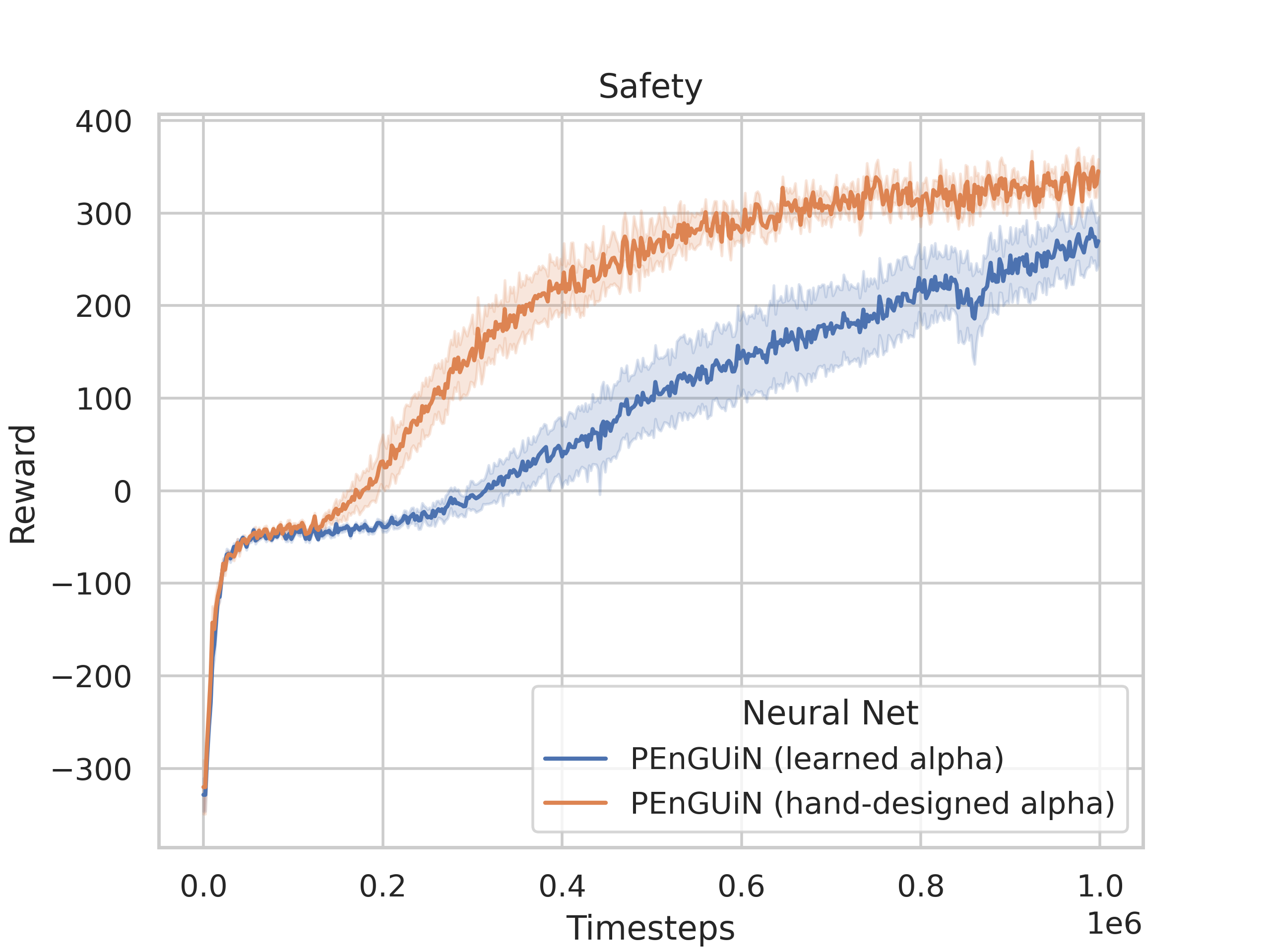}
    \caption{\small{An example of using domain knowledge to hand-design $\alpha$} }
    \label{fig:hard_coded_alpha}
    \vspace{-20pt}
\end{wrapfigure}
Figure \ref{fig:alpha_heatmap} reveals that PEnGUiN indeed learns to modulate equivariance spatially.  The heatmap shows lower $\alpha$ values concentrated in the upper-right quadrant, corresponding to the safety region. This suggests that PEnGUiN successfully identifies the region where equivariance is broken and reduces its reliance on equivariant updates in that area, while maintaining higher equivariance in the symmetric regions of the environment.

Finally, we experiment with using a hand-designed value for $\alpha$. If an engineer can identify the symmetries and asymmetries in a scenario, they may encode that into the neural network, improving the inductive bias of the model. For this experiment, we use the simple tag safety environment. We set $\alpha = 0$ when  $\vx > \textbf{0}$ (i.e. where the safety region violates equivariance and then set  $\alpha=1$ for the remaining locations. In figure \ref{fig:hard_coded_alpha}, we see the results of this simple experiment. Hand designing $\alpha$, in this case, does indeed seem to improve sample efficiency. This may not always be the case, there are many environments where hand designing $\alpha$ may be nontrivial, especially when one considers that $\alpha$ is used for all layers, and the optimal $\alpha$ may depend on the layer. 

\subsection{Experiments on Highway Environment}

\subsubsection{Environment Setup}

To assess PEnGUiN's performance in more complex and realistic scenarios, we evaluated it on the highway-env benchmark \cite{highway-env}, a suite of environments for autonomous driving. We focused on two challenging environments: Racetrack and Roundabout. \textbf{Racetrack:} In this environment, the agent must navigate a closed racetrack, following the track's curvature while maintaining speed and avoiding collisions with other vehicles. \textbf{Roundabout:} This scenario requires agents to navigate a roundabout intersection, performing lane changes and speed adjustments to efficiently pass through the roundabout while avoiding collisions.

\begin{wrapfigure}{hl}{0.5\textwidth}
    \includegraphics[width=0.5\textwidth]{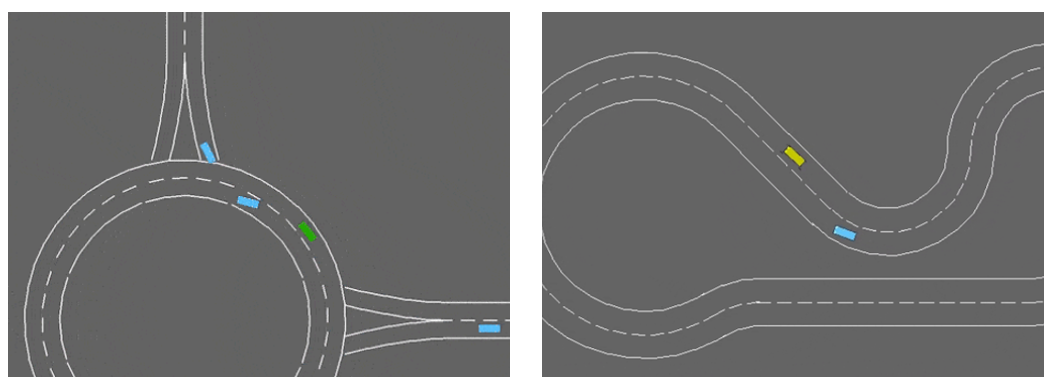}
    \caption{\small{Vizualization of roundabout and racetrack scenario }}
    \label{fig:highway_imgs}
\end{wrapfigure}

These environments utilize a more sophisticated bicycle dynamics model for vehicle motion, introducing non-linear dynamics and requiring precise control over steering and throttle actions.  Furthermore, the constraint of staying within the road boundaries and lanes naturally introduces a form of regional equivariance, as symmetry is broken at the road edges.

We maintained consistent implementation details with the MPE experiments, using RLLib PPO and comparing PEnGUiN against the same set of baselines (EGNN, E2GN2, and GNN).

\subsubsection{Results and Discussion (highway-env)}
\begin{figure}[htbp]
    \begin{subfigure}[b]{0.45\columnwidth}
        \centering
        \includegraphics[width=0.9\columnwidth]{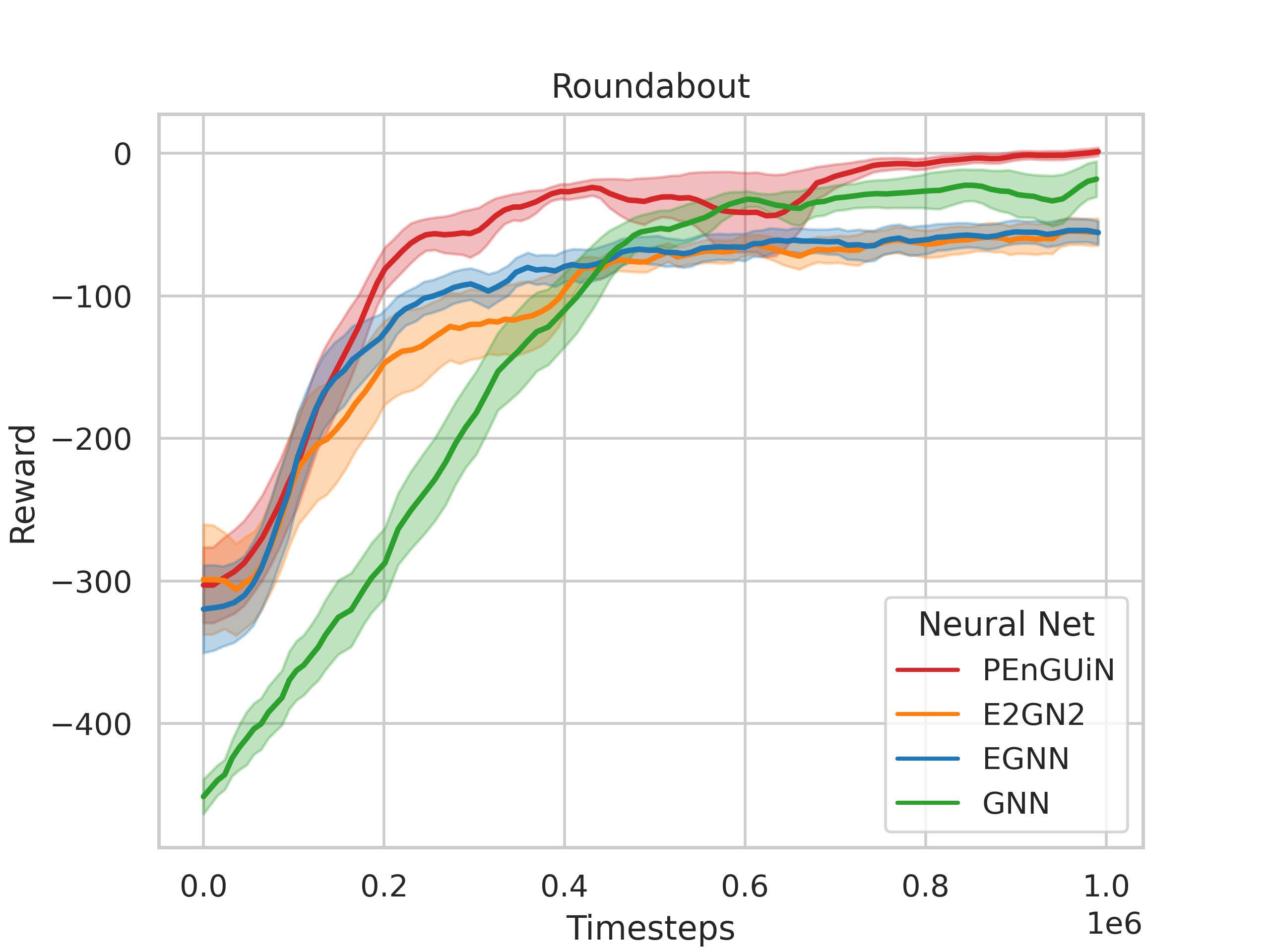}
    \end{subfigure}
    \begin{subfigure}[b]{0.45\columnwidth}
        \centering
        \includegraphics[width=0.9\columnwidth]{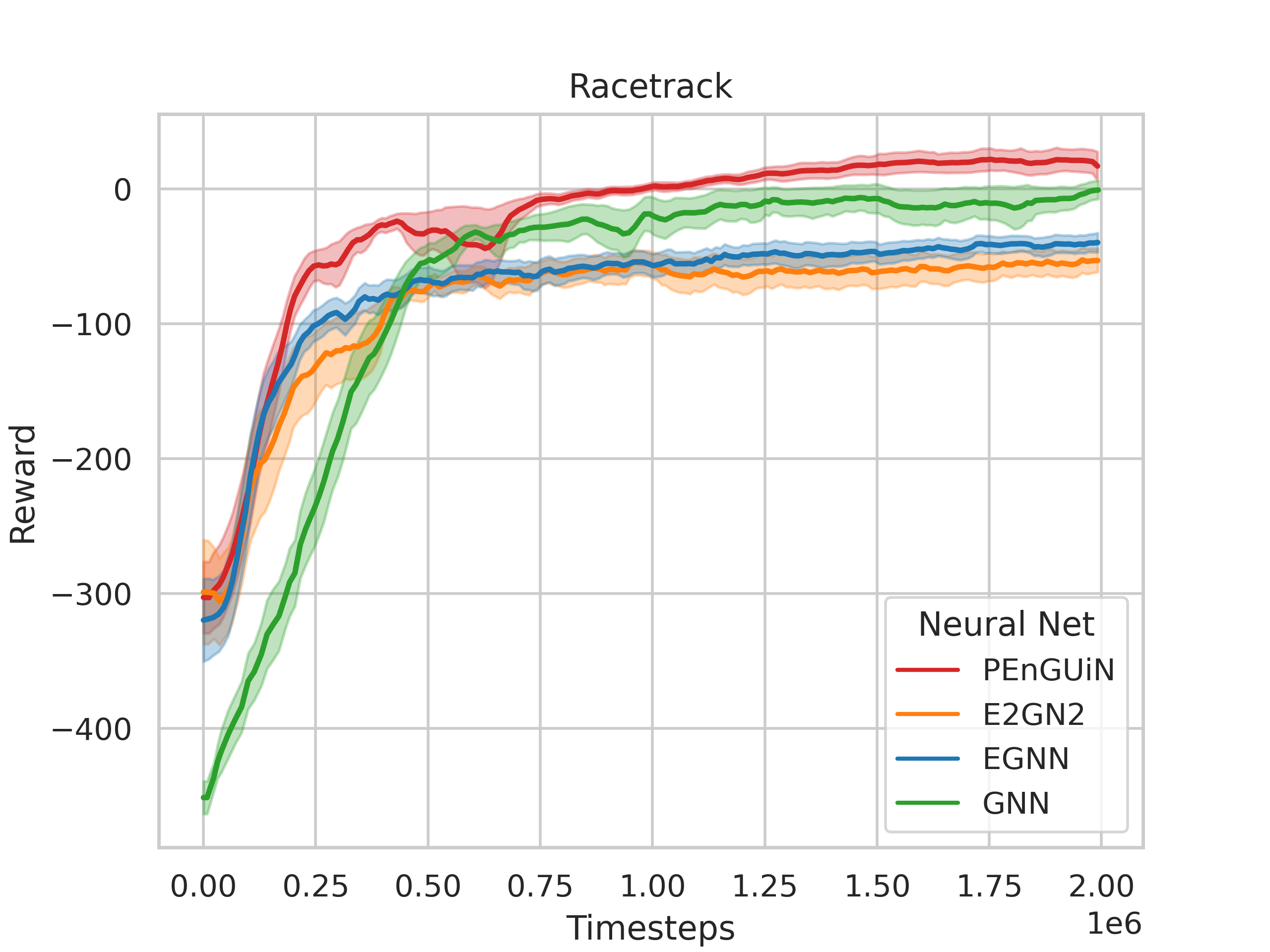}
    \end{subfigure}
    \caption{\small{Learning curves on highway-env \textit{racetrack} and \textit{roundabout} environments. Results are averaged over multiple seeds with shaded regions indicating standard error. PEnGUiN consistently outperforms EGNN, E2GN2, and GNN, demonstrating its effectiveness in more complex environments with non-linear dynamics and regional constraints.}}
    \label{fig:highway_learning_curves}
\end{figure}

Figure \ref{fig:highway_learning_curves} presents the learning curves for the highway-env racetrack and roundabout scenarios.

The results in Figure \ref{fig:highway_learning_curves} demonstrate that PEnGUiN consistently outperforms all baselines in both highway-env scenarios. PEnGUiN achieves higher rewards and exhibits faster convergence compared to EGNN, E2GN2, and GNN. This indicates that PEnGUiN's ability to adapt to partial equivariance is beneficial even in environments with more complex, non-linear dynamics and regional constraints, where full equivariance might be a suboptimal inductive bias.



\section{Conclusion}

This paper introduced Partially Equivariant Graph Neural Networks (PEnGUiN), a novel architecture for Multi-Agent Reinforcement Learning (MARL) that addresses the limitations of existing fully equivariant models in real-world, partially symmetric environments. Unlike traditional Equivariant Graph Neural Networks (EGNNs) that assume full symmetry, PEnGUiN learns to blend equivariant and non-equivariant updates, controlled by a learnable parameter. This allows it to adapt to various types of partial equivariance, including subgroup, feature-wise, subspace, and approximate equivariance, which we formally defined and categorized.

We theoretically demonstrated that PEnGUiN encompasses both fully equivariant (EGNN) and non-equivariant (GNN) representations as special cases, providing a unified and flexible framework.  Extensive experiments on modified Multi-Particle Environments (MPE) and the more complex highway-env benchmark showed that PEnGUiN consistently outperforms both EGNNs and standard GNNs in scenarios with various asymmetries, demonstrating improved sample efficiency and robustness.  Furthermore, visualizations of the Equivariance Estimator explored PEnGUiN's ability to identify and exploit regions and features where equivariance holds and where it is violated. PEnGUiN expands the applicability of equivariant graph neural networks to real-world MARL by handling partial symmetries, common in scenarios like robotics, autonomous driving, and multi-agent systems with sensor biases or external forces. By learning to navigate the complexities of partial symmetries, PEnGUiN represents a step towards realizing safe and dependable multi-agent robotic systems in the real world.

\appendix

\section{Appendix A: Proofs}
For convenience we rewrite the equations for a single GNN update using node $i$ embeddings $\vh_i \in \sR^h$ and intermediate messages between node $i$ and $j$: $\vn_{ij} \in \sR^m$. where $\phi_{h}$ and $\phi_{e}$ are the MLPs. We will use superscripts $l$ below to denote the layer.
\[
\vn_{ij} =\phi_{e}\lp\vh_{i}^{l}, \vh_{j}^{l}\ep, \quad \vn_{i} =\sum_{j \neq i} \vn_{ij}, \quad \vh_{i}^{l+1} =\phi_{h}\lp\vh_{i}^l, \vn_{i}\ep   
\]

\subsection{Proof PEnGUiN embeds an E2GN2}
Setting $\alpha = 1$ in the PEnGUiN equations directly yields the EGNN equations. For clarity we will rewrite the partially equivariant node and coordinate update equations and how it changes when $\alpha=1$:
$$ \vm_{i}^{l} = \alpha \sum_{j \neq i} \vm_{ij}^l + (1-\alpha) \sum_{j \neq i} \vn_{ij}^l = \alpha \sum_{j \neq i} \vm_{ij}^l$$
$$\vu_{i}^{l+1} = \alpha  \vu_{i,eq}^l  + (1-\alpha) \vu_i^p = \vu_{i,eq}^l$$

Then when $alpha=1$ the update equations become:
$$\vu_{i,eq}^{l} = \vu_{i}^{l} \phi_{e}(\vm_i^l) + \sum_{j \neq i} \left(\vu_{i}^{l} -\vu_{j}^{l}\right) \phi_{u} \left(\vm_{ij}^l\right)) $$
$$\vm_{i}^{l} =  \sum_{j \neq i} \vm_{ij}^l \quad \vh_{i}^{l+1} = \phi_{h}\left(\vh_{i}^{l}, \vm_{i}^{l}\right) $$
These are precisely the update equations for an E2GN2 layer.


\subsection{Proof of GNN equivalence}
 
Proof  PEnGUiN is equivalent to a GNN when $\alpha=0$. Recall the node embeddings $\vh_i \in \sR^h$ and the coordinate embeddings $\vu_i \in \sR^n$  For clarity we will rewrite the partially equivariant node and coordinate updates (the equations with $\alpha$), and how it changes when $\alpha=0$:
$$
 \vm_{i}^{l} = \alpha \sum_{j \neq i} \vm_{ij}^l + (1-\alpha) \sum_{j \neq i} \vn_{ij}^l = \sum_{j \neq i} \vn_{ij}^l
$$
$$\vu_{i}^{l+1} = \alpha  \vu_{i,eq}^l  + (1-\alpha) \vu_i^p = \vu_i^p$$

Thus far this means that our output $\vh_i$ will be purely using the GNN update message $n_{ij}$. Next we will note that we can rewrite  $\vh_{i}^{l+1}, \vu_i^p$ as $\vh_{i,0:h}^l, \vh_{i,h:h+n}^l$ (essentially this is simply renaming notation. In the main text, we used $\vu_i^p$ to aid in clarity). We use this renaming to represent that $\vh_{i,0:h}^l$ contains the first $h$ elements of the output from $\phi_h$, and $\vh_{i,h:h+n}^l$ is the remaining $n$ elements. Thus the final node update for this layer becomes:
$\vh_{i,0:h}^l, \vh_{i,h:h+n}^l =\phi_{h}\left(\vh_{i}^{l}, \vm_{i}^{l}\right)$ 

To ensure this is equivalent to a GNN, we now look at the next layer in the network. We now see that the next layer becomes:

$$
\vn_{ij} =\phi_{n}\lp\vh_{i}^{l+1}, \vh_{j}^{l+1},\vu_{i}^{l+1}, \vu_{j}^{l+1}\ep 
= \phi_{n}\lp 
\vh_{i,0:L-2}^{l+1}, \vh_{j,0:L-2}^{l+1},\vh_{j,L-2:L}^{l+1}, \vh_{j,L-2:L}^{l+1}
\ep
$$
This is equivalent to a standard GNN messgae update which is: $\phi_{n}\lp \vh_i, vh_j \ep$ The only difference is that we explicitely separate (then later concatenate) the last $n$ elements of $\vh$
The remainder of the equations of PEnGUiN for layer $l+1$ (with $\alpha=0$) will be:
$ \vm_{i}^{l+2} = \sum_{j \neq i} \vn_{ij}^{l+1}$, with the final node update:
$\vh_{i,0:h}^{l+2}, \vh_{i,h:h+n}^l =\phi_{h}\left(\vh_{i}^{l+1}, \vm_{i}^{l+1}\right)$ 
which is equivalent to a GNN







\bibliography{main}
\bibliographystyle{rlj}

\beginSupplementaryMaterials

\section{Additional Training Details} \label{appndx:hyper}

\begin{table}[h]
\centering
\begin{tabular}{cc}
\toprule
Hyperparameters      & value  \\
\midrule
train batch size            & 2000 \\
mini-batch size            & 1000 \\
PPO clip                     & 0.2 \\
learning rate            & 30e-5       \\
num SGD iterations      & 10       \\
gamma                     & 0.99 \\
lambda              & 0.95       \\

\bottomrule
\end{tabular}
\caption{hyperparameters for MPE}
\end{table}

\begin{table}[h]
\centering
\begin{tabular}{cc}
\toprule
Hyperparameters      & value  \\
\midrule
train batch size            & EpisodeLength*16 \\
mini-batch size            & EpisodeLength*4 \\
PPO clip                     & 0.2 \\
learning rate            & 45e-5       \\
num SGD iterations      & 10       \\
gamma                     & 0.99 \\
lambda              & 0.95       \\
\bottomrule
\end{tabular}
\caption{PPO Common Hyperparameters for Highway-env}
\end{table}

 All MLPs in the GNNs use 2 layers with a width of 32. For all GNN structures we use separate networks for the policy and value functions.

\textbf{Graph Structure and Inputs} The graph structure for MPE environments is set as a complete graph. For MPE environments the input invariant feature for each node $\vh_i^0$ is the id (pursuer, evader, or landmark).  For MPE there is also a velocity feature, which we incoporate following the procedure described in \citep{egnn}. For the Highway-env, we incorporated the 

\textbf{Graph Outputs for Value function and Policy} We followed the design choices in \citep{E2GN2} for the action space and value function design: the value function output comes from the invariant component of the agent's node of final layer of the EGNN/E2GN2. For MPE the actions are (partially) equivariant, so we use the outputs of the coordinate embeddings. For highway env, the actions are (partially) invariant, so we use the outputs of $\vh_i$ for the policy output

\textbf{Experiment design} For the MPE simple tag environment we added a hard-coded evader agent. This agent computed the force as $force = - x_i + \frac{0.3}{N} \sum_j (x_j-x_i)/ || (x_j - x_i ||^2 )$ where $x_i$ is the evader's position, and $j$ corresponds to each pursuer. Essentially the agent tries to move away from the pursuers, and to stay close to the center of the world (to avoid going off to infinity). The force from the pursuers is limited to not be larger than 2. The final force is normalized and divided by 3.  
As described in the main text we have 3 modifications to MPE. The bias modification is to add a value of 0.3 in the x direction to each landmark and evader. The safety scenario gives a negative reward when the agent is in the region $x, y > 0$. In simple tag this reward is -15, for spread it is -5. Note that for spread the agents are not initialized within the safety region. Finally the decoy environment consists of having the objective (landmarks/evader) being static. For spread there are three decoy landmarks, with the actual landmarks at the xy positions: (1.5,.9),(-.9,0.),(-.5,-.5). The tag environment has one decoy with the actual evader at (.75,.75).

\end{document}

%% file: math_commands.tex
\usepackage{amsmath,amsfonts,bm}

\def\eqref#1{equation~\ref{#1}}

\def\1{\bm{1}}

\def\vh{{\bm{h}}}

\def\vm{{\bm{m}}}
\def\vn{{\bm{n}}}

\def\vu{{\bm{u}}}

\def\vx{{\bm{x}}}

\DeclareMathAlphabet{\mathsfit}{\encodingdefault}{\sfdefault}{m}{sl}
\SetMathAlphabet{\mathsfit}{bold}{\encodingdefault}{\sfdefault}{bx}{n}

\def\sR{{\mathbb{R}}}










%% file: main.bbl
\begin{thebibliography}{28}
\providecommand{\natexlab}[1]{#1}
\providecommand{\url}[1]{\texttt{#1}}
\expandafter\ifx\csname urlstyle\endcsname\relax
  \providecommand{\doi}[1]{DOI: #1}\else
  \providecommand{\doi}{DOI: \begingroup \urlstyle{rm}\Url}\fi

\bibitem[Brandstetter et~al.(2022)Brandstetter, Hesselink, van~der Pol, Bekkers, and Welling]{segnn}
Johannes Brandstetter, Rob Hesselink, Elise van~der Pol, Erik~J. Bekkers, and Max Welling.
\newblock Geometric and physical quantities improve {E(3)} equivariant message passing.
\newblock In \emph{The Tenth International Conference on Learning Representations, {ICLR} 2022, Virtual Event, April 25-29, 2022}. OpenReview.net, 2022.
\newblock URL \url{https://openreview.net/forum?id=\_xwr8gOBeV1}.

\bibitem[Chen \& Zhang(2024)Chen and Zhang]{e3_coop}
Dingyang Chen and Qi~Zhang.
\newblock ${\rm e}(3)$-equivariant actor-critic methods for cooperative multi-agent reinforcement learning, 2024.

\bibitem[Chen et~al.(2023)Chen, Han, Sun, and Huang]{subequiv_rl}
Runfa Chen, Jiaqi Han, Fuchun Sun, and Wenbing Huang.
\newblock Subequivariant {Graph} {Reinforcement} {Learning} in {3D} {Environments}.
\newblock In \emph{Proceedings of the 40th {International} {Conference} on {Machine} {Learning}}, pp.\  4545--4565. PMLR, July 2023.
\newblock URL \url{https://proceedings.mlr.press/v202/chen23i.html}.
\newblock ISSN: 2640-3498.

\bibitem[Finzi et~al.(2021{\natexlab{a}})Finzi, Benton, and Wilson]{finzi_residual_2021}
Marc Finzi, Gregory Benton, and Andrew~Gordon Wilson.
\newblock Residual {Pathway} {Priors} for {Soft} {Equivariance} {Constraints}, December 2021{\natexlab{a}}.
\newblock URL \url{http://arxiv.org/abs/2112.01388}.
\newblock arXiv:2112.01388 [cs].

\bibitem[Finzi et~al.(2021{\natexlab{b}})Finzi, Welling, and Wilson]{Finzi2021}
Marc Finzi, Max Welling, and Andrew~Gordon Wilson.
\newblock A practical method for constructing equivariant multilayer perceptrons for arbitrary matrix groups, 2021{\natexlab{b}}.

\bibitem[Geiger \& Smidt(2022)Geiger and Smidt]{e3nn}
Mario Geiger and Tess Smidt.
\newblock e3nn: Euclidean neural networks, 7 2022.
\newblock URL \url{https://arxiv.org/abs/2207.09453v1}.

\bibitem[Hofgard et~al.(2024)Hofgard, Wang, Walters, and Smidt]{relax_eq_gnn}
Elyssa Hofgard, Rui Wang, Robin Walters, and Tess Smidt.
\newblock Relaxed {Equivariant} {Graph} {Neural} {Networks}, December 2024.
\newblock URL \url{http://arxiv.org/abs/2407.20471}.
\newblock arXiv:2407.20471 [cs].

\bibitem[Huang et~al.(2023)Huang, Levie, and Villar]{villar_approx_gnn}
Ningyuan~Teresa Huang, Ron Levie, and Soledad Villar.
\newblock Approximately {Equivariant} {Graph} {Networks}.
\newblock November 2023.
\newblock URL \url{https://openreview.net/forum?id=5aeyKAZr0L}.

\bibitem[Leurent(2018)]{highway-env}
Edouard Leurent.
\newblock An environment for autonomous driving decision-making.
\newblock \url{https://github.com/eleurent/highway-env}, 2018.

\bibitem[Liang et~al.(2018)Liang, Liaw, Nishihara, Moritz, Fox, Goldberg, Gonzalez, Jordan, and Stoica]{rllib}
Eric Liang, Richard Liaw, Robert Nishihara, Philipp Moritz, Roy Fox, Ken Goldberg, Joseph~E. Gonzalez, Michael~I. Jordan, and Ion Stoica.
\newblock {RLlib}: Abstractions for distributed reinforcement learning.
\newblock In \emph{International Conference on Machine Learning ({ICML})}, 2018.

\bibitem[Littman(1994)]{littman1994markov}
Michael~L. Littman.
\newblock \emph{Markov games as a framework for multi-agent reinforcement learning}.
\newblock Morgan Kaufmann, San Francisco (CA), 1994.
\newblock ISBN 978-1-55860-335-6.
\newblock \doi{https://doi.org/10.1016/B978-1-55860-335-6.50027-1}.
\newblock URL \url{https://www.sciencedirect.com/science/article/pii/B9781558603356500271}.

\bibitem[Lowe et~al.(2017)Lowe, WU, Tamar, Harb, Pieter~Abbeel, and Mordatch]{maddpg}
Ryan Lowe, YI~WU, Aviv Tamar, Jean Harb, OpenAI Pieter~Abbeel, and Igor Mordatch.
\newblock Multi-agent actor-critic for mixed cooperative-competitive environments.
\newblock In I.~Guyon, U.~Von Luxburg, S.~Bengio, H.~Wallach, R.~Fergus, S.~Vishwanathan, and R.~Garnett (eds.), \emph{Advances in Neural Information Processing Systems}, volume~30. Curran Associates, Inc., 2017.
\newblock URL \url{https://proceedings.neurips.cc/paper_files/paper/2017/file/68a9750337a418a86fe06c1991a1d64c-Paper.pdf}.

\bibitem[McClellan et~al.(2024)McClellan, Haghani, Winder, Huang, and Tokekar]{E2GN2}
Josh McClellan, Naveed Haghani, John Winder, Furong Huang, and Pratap Tokekar.
\newblock Boosting sample efficiency and generalization in multi-agent reinforcement learning via equivariance.
\newblock In A.~Globerson, L.~Mackey, D.~Belgrave, A.~Fan, U.~Paquet, J.~Tomczak, and C.~Zhang (eds.), \emph{Advances in Neural Information Processing Systems}, volume~37, pp.\  41132--41156. Curran Associates, Inc., 2024.
\newblock URL \url{https://proceedings.neurips.cc/paper_files/paper/2024/file/4830a9b95a2f63fc4b3fe09abc18f045-Paper-Conference.pdf}.

\bibitem[McNeela(2024)]{almost_eq_lieConv}
Daniel McNeela.
\newblock Almost {Equivariance} via {Lie} {Algebra} {Convolutions}, June 2024.
\newblock URL \url{http://arxiv.org/abs/2310.13164}.
\newblock arXiv:2310.13164 [cs].

\bibitem[Ouderaa et~al.(2022)Ouderaa, Romero, and Wilk]{relax_eq_filters}
Tycho F. A. van~der Ouderaa, David~W. Romero, and Mark van~der Wilk.
\newblock Relaxing {Equivariance} {Constraints} with {Non}-stationary {Continuous} {Filters}, November 2022.
\newblock URL \url{http://arxiv.org/abs/2204.07178}.
\newblock arXiv:2204.07178 [cs].

\bibitem[Park et~al.(2024)Park, Bhatt, Zeng, Wong, Koppel, Ganesh, and Walters]{approx_eq_rl}
Jung~Yeon Park, Sujay Bhatt, Sihan Zeng, Lawson L.~S. Wong, Alec Koppel, Sumitra Ganesh, and Robin Walters.
\newblock Approximate {Equivariance} in {Reinforcement} {Learning}, November 2024.
\newblock URL \url{http://arxiv.org/abs/2411.04225}.
\newblock arXiv:2411.04225 [cs].

\bibitem[Petrache \& Trivedi()Petrache and Trivedi]{approx_gen_group_eq}
Mircea Petrache and Shubhendu Trivedi.
\newblock Approximation-{Generalization} {Trade}-offs under ({Approximate}) {Group} {Equivariance}.

\bibitem[Pol et~al.(2021)Pol, Hoof, Deltalab, Oliehoek, and Welling]{eqmmdp}
Elise Van~Der Pol, Herke~Van Hoof, Uva-Bosch Deltalab, Frans~A Oliehoek, and Max Welling.
\newblock Multi-agent mdp homomorphic networks, 10 2021.
\newblock URL \url{https://arxiv.org/abs/2110.04495v2}.

\bibitem[Samudre et~al.(2024)Samudre, Petrache, Nord, and Trivedi]{matrix_cnn_approx}
Ashwin Samudre, Mircea Petrache, Brian~D. Nord, and Shubhendu Trivedi.
\newblock Symmetry-{Based} {Structured} {Matrices} for {Efficient} {Approximately} {Equivariant} {Networks}, September 2024.
\newblock URL \url{http://arxiv.org/abs/2409.11772}.
\newblock arXiv:2409.11772 [stat].

\bibitem[Satorras et~al.(2021)Satorras, Hoogeboom, and Welling]{egnn}
V\'{\i}ctor~Garcia Satorras, Emiel Hoogeboom, and Max Welling.
\newblock E(n) equivariant graph neural networks.
\newblock In \emph{Proceedings of the 38th International Conference on Machine Learning}, volume 139 of \emph{Proceedings of Machine Learning Research}, pp.\  9323--9332. PMLR, 18--24 Jul 2021.
\newblock URL \url{https://proceedings.mlr.press/v139/satorras21a.html}.

\bibitem[Schulman et~al.(2017)Schulman, Wolski, Dhariwal, Radford, and Klimov]{ppo}
John Schulman, Filip Wolski, Prafulla Dhariwal, Alec Radford, and Oleg Klimov.
\newblock Proximal policy optimization algorithms.
\newblock \emph{CoRR}, abs/1707.06347, 2017.
\newblock URL \url{http://arxiv.org/abs/1707.06347}.

\bibitem[van~der Pol et~al.(2020)van~der Pol, Worrall, van Hoof, Oliehoek, and Welling]{eqmdp}
Elise van~der Pol, Daniel~E. Worrall, Herke van Hoof, Frans~A. Oliehoek, and Max Welling.
\newblock Mdp homomorphic networks: Group symmetries in reinforcement learning.
\newblock \emph{Advances in Neural Information Processing Systems}, 2020-December, 6 2020.
\newblock ISSN 10495258.
\newblock URL \url{https://arxiv.org/abs/2006.16908v2}.

\bibitem[Wang et~al.(2022{\natexlab{a}})Wang, Park, Sortur, Wong, Walters, and Platt]{eff_latent_sym}
Dian Wang, Jung~Yeon Park, Neel Sortur, Lawson L.~S. Wong, Robin Walters, and Robert Platt.
\newblock The {Surprising} {Effectiveness} of {Equivariant} {Models} in {Domains} with {Latent} {Symmetry}.
\newblock September 2022{\natexlab{a}}.
\newblock URL \url{https://openreview.net/forum?id=P4MUGRM4Acu}.

\bibitem[Wang et~al.(2022{\natexlab{b}})Wang, Walters, and Platt]{Wang2022}
Dian Wang, Robin Walters, and Robert Platt.
\newblock $\mathrm\{SO\}(2)$-equivariant reinforcement learning, 3 2022{\natexlab{b}}.
\newblock URL \url{https://arxiv.org/abs/2203.04439v1}.

\bibitem[Wang et~al.(2023)Wang, Zhu, Park, Jia, Su, Platt, and Walters]{extrinsic_eq_theory}
Dian Wang, Xupeng Zhu, Jung~Yeon Park, Mingxi Jia, Guanang Su, Robert Platt, and Robin Walters.
\newblock A {General} {Theory} of {Correct}, {Incorrect}, and {Extrinsic} {Equivariance}.
\newblock November 2023.
\newblock URL \url{https://openreview.net/forum?id=2FMJtNDLeE&noteId=LtiReb7xPp}.

\bibitem[Wang et~al.(2022{\natexlab{c}})Wang, Walters, and Yu]{approx_eq_def}
Rui Wang, Robin Walters, and Rose Yu.
\newblock Approximately {Equivariant} {Networks} for {Imperfectly} {Symmetric} {Dynamics}, June 2022{\natexlab{c}}.
\newblock URL \url{http://arxiv.org/abs/2201.11969}.
\newblock arXiv:2201.11969 [cs].

\bibitem[Wang et~al.(2024)Wang, Hofgard, Gao, Walters, and Smidt]{relax_groupConv}
Rui Wang, Elyssa Hofgard, Han Gao, Robin Walters, and Tess~E. Smidt.
\newblock Discovering {Symmetry} {Breaking} in {Physical} {Systems} with {Relaxed} {Group} {Convolution}, June 2024.
\newblock URL \url{http://arxiv.org/abs/2310.02299}.
\newblock arXiv:2310.02299 [cs].

\bibitem[Yu et~al.(2024)Yu, Shi, Feng, Tian, Li, Liao, and Wu]{rl_symloss}
Xin Yu, Rongye Shi, Pu~Feng, Yongkai Tian, Simin Li, Shuhao Liao, and Wenjun Wu.
\newblock Leveraging partial symmetry for multi-agent reinforcement learning.
\newblock \emph{Proceedings of the AAAI Conference on Artificial Intelligence}, 38\penalty0 (16):\penalty0 17583--17590, Mar. 2024.
\newblock \doi{10.1609/aaai.v38i16.29709}.
\newblock URL \url{https://ojs.aaai.org/index.php/AAAI/article/view/29709}.

\end{thebibliography}
